%% file: root.tex
\documentclass[letterpaper, 10 pt, conference]{ieeeconf}  

\IEEEoverridecommandlockouts                              

\usepackage{hyperref}
\hypersetup{
    colorlinks=true,
    linkcolor=blue,
    citecolor=blue,
    urlcolor=blue,
    pdfborder={0 0 0}  
}
\usepackage{adjustbox}
\usepackage{booktabs}
\usepackage{subcaption}
\usepackage{amsmath}
\usepackage{amssymb}
\usepackage{cite}
\usepackage{multirow}
\usepackage{float}
\usepackage[table]{xcolor}
\let\labelindent\relax
\usepackage{enumitem}

\title{\LARGE \bf
Diffusion-Based Generative Models for 3D Occupancy Prediction in Autonomous Driving 
}

\author{Yunshen Wang$^{1,2,\star}$, Yicheng Liu$^{1,\star}$, Tianyuan Yuan$^{ 1,\star}$, Yingshi Liang$^{2}$, \\Xiuyu Yang$^{1}$, Honggang Zhang$^{2}$, Hang Zhao$^{1,\dag}$
\thanks{$^{\star}$equal contribution}
\thanks{
$^{1}$ Institute for Interdisciplinary Information Sciences, Tsinghua University.
$^{2}$ Beijing University of Posts and Telecommunications. }
\thanks{\dag Corresponding to: hangzhao@mail.tsinghua.edu.cn}
}

\begin{document}

\maketitle
\thispagestyle{empty}
\pagestyle{empty}

\begin{abstract}

Accurately predicting 3D occupancy grids from visual inputs is critical for autonomous driving, but current discriminative methods struggle with noisy data, incomplete observations, and the complex structures inherent in 3D scenes. In this work, we reframe 3D occupancy prediction as a generative modeling task using diffusion models, which learn the underlying data distribution and incorporate 3D scene priors. This approach enhances prediction consistency, noise robustness, and better handles the intricacies of 3D spatial structures. Our extensive experiments show that diffusion-based generative models outperform state-of-the-art discriminative approaches, delivering more realistic and accurate occupancy predictions, especially in occluded or low-visibility regions. Moreover, the improved predictions significantly benefit downstream planning tasks, highlighting the practical advantages of our method for real-world autonomous driving applications.

\end{abstract}

\input{figs/teaser}

\section{INTRODUCTION}
Vision-based 3D occupancy prediction is a task focused on estimating the semantic labels and occupancy states of each voxel in a scene from visual inputs, helping autonomous vehicles perceive their 3D environment with centimeter-level precision. Despite the recent progress in dataset, models and benchmarks~\cite{liao2022kitti,semantickitti,occ3d}, accurately predicting the 3D occupancy grids is still a highly challenging task.

Recent approaches in 3D occupancy prediction, which we refer to as discriminative methods, directly learn a mapping from images to occupancy grids and have become the de facto choice.
However, there are inherent challenges in solving the 3D occupancy prediction task using discriminative methods: 1) The unique nature of predicting occupancy from visual inputs—such as the complex 3D structures, the intricate relationships between 3D labels and the existence of multimodal distributions—makes this task distinct and more challenging compared to other supervised learning problems. However, discriminative methods directly learn the mapping from images to occupancy, rather than estimating the underlying distributions, which prevents them from incorporating prior knowledge of 3D scenes. This often leads to unrealistic and inconsistent results (refer to the black and orange circle in Fig.~\ref{fig:teaser}). Moreover, these methods fail to capture the multimodal nature of the distributions, increasing the burden on downstream tasks. 2) Obtaining perfect occupancy map labels is almost infeasible. Existing benchmarks like KITTI360 [26] and Occ3D [39] rely on LiDAR scans from one or multiple traversals to create mesh reconstructions of scenes. However, partial observations and sensor noise lead to imperfect labels, which hinder effective model learning.

Given these challenges, we believe that modeling occupancy prediction as a generative modeling problem offers a promising solution. By directly modeling the occupancy data distribution and performing conditional sampling, generative modeling captures the prior knowledge of complex 3D structures and semantics while naturally considering the inherent multi-modality of the occupancy prediction task. Furthermore, mainstream generative models, such as Diffusion Models exhibit inherent robustness to noise, which can mitigate the issue of harmful noise in occupancy labels. Moreover, generative models aim to model the underlying distribution of 3D world occupancy, rather than optimizing a direct mapping from images to occupancy. This results in better generalization across scenarios with varying sensor setups, such as input images that differ from the training set or lack certain information.

Motivated by this, we explore how to leverage diffusion models
for occupancy prediction, including investigating how to perform generative modeling and conditional sampling, and explore other interesting properties. Our extensive experiments demonstrate that generative modeling for occupancy prediction offers a series of powerful advantages over discriminative modeling. First, directly modeling the data distribution introduces a strong prior for 3D scene occupancy, enhancing the model’s perceptual capabilities, resulting in more realistic, consistent, and accurate outcomes. For regions with high uncertainty, such as those with insufficient observation, occlusion, or high levels of noise, our model exhibits superior perceptual capabilities. Such holistic, uncertainty-aware, and multimodal-considerate perception results also better support downstream planning tasks, as shown by our experimental results. Our key contributions are summarized as follows:
\begin{itemize}
\item We frame occupancy prediction as a process involving generative modeling followed by conditional sampling, from which we summarized four appealing properties compare to discriminative counterpart.
\item We explore five key design aspects of utilizing conditional generative modeling for the occupancy prediction task.
\item Through extensive experiments, we demonstrate that incorporating diffusion models can significantly improve the performance of occupancy prediction. The occupancy features generated by our method also benefit downstream planning tasks.
\end{itemize}

\section{Formulation}
We formulate the task of occupancy prediction using diffusion models \cite{ddpm,ddim,hoogeboom2021argmax,austin2021structured}, which can express complex multimodal distributions and exhibit robustness to noise. Below, we provide an overview of diffusion models and their adaptation for occupancy prediction.

\subsection{Diffusion Models}
Diffusion models progressively add noise to data and reverse this process to generate samples. For continuous data, the forward process adds Gaussian noise at each step:
\begin{equation}
    q(\mathbf{x}_t|\mathbf{x}_{t-1}) = \mathcal{N}(\mathbf{x}_t; \sqrt{1-\beta_t}\mathbf{x}_{t-1}, \beta_t \mathbf{I}),
    \label{eq:gaussian_forward}
\end{equation}
where \(\beta_t\) controls the noise level at step \(t\). The reverse process, parameterized by \(\theta\), denoises the data:
\begin{equation}
    p_{\theta}(\mathbf{x}_{t-1}|\mathbf{x}_t) = \mathcal{N}(\mathbf{x}_{t-1}; \mu_\theta(\mathbf{x}_t, t), \sigma^2_\theta(t) \mathbf{I}),
    \label{eq:gaussian_reverse}
\end{equation}

For discrete data, diffusion models~\cite{hoogeboom2021argmax,austin2021structured} replace Gaussian noise with a discrete corruption process, using a transition matrix \(\mathbf{Q}_t\)\cite{austin2021structured}. For discrete variables \(\mathbf{x}_t, \mathbf{x}_{t-1} \in \{1, \ldots, K\}\), the forward process becomes:
\begin{equation}
    q(\mathbf{x}_t|\mathbf{x}_{t-1}) = \text{Cat}(\mathbf{x}_t; \mathbf{p} = \mathbf{x}_{t-1}\mathbf{Q}_t),
    \label{eq:discrete_forward}
\end{equation}
The reverse process then predicts the discrete state transitions:
\begin{equation}
    p_{\theta}(\mathbf{x}_{t-1}|\mathbf{x}_t) = \text{Cat}(\mathbf{x}_{t-1}; \mathbf{p} = f_\theta(\mathbf{x}_t, t)).
    \label{eq:discrete_reverse}
\end{equation}

\subsection{Training Diffusion Models}
Starting from a sample \(\mathbf{x}_0\), the marginal at step \(t\) is \(q(\mathbf{x}_t|\mathbf{x}_0)\),
with the posterior:
\begin{equation}
    q(\mathbf{x}_{t-1}|\mathbf{x}_t, \mathbf{x}_0) = \frac{q(\mathbf{x}_t|\mathbf{x}_{t-1})q(\mathbf{x}_{t-1}|\mathbf{x}_0)}{q(\mathbf{x}_t|\mathbf{x}_0)}.
    \label{eq:q_post}
\end{equation}

The reverse process is optimized by minimizing the KL divergence between the forward process and the predicted reverse process:
\begin{equation}
    \text{loss} = D_{KL}\left( q(\mathbf{x}_{t-1}|\mathbf{x}_t, \mathbf{x}_0) \,\|\, p_{\theta}(\mathbf{x}_{t-1}|\mathbf{x}_t) \right).
\end{equation}

\subsection{Occupancy Prediction as Conditional Generation}
We adapt diffusion models to predict occupancy by modifying the output \(\mathbf{x}\) to represent occupancy grids and conditioning the reverse process on input multi-view images \(\mathbf{C}\). This modifies Eq.~\eqref{eq:gaussian_reverse} to:
\begin{equation}
    p_{\theta}(\mathbf{x}_{t-1}|\mathbf{x}_t, \mathbf{C}) = \mathcal{N}(\mathbf{x}_{t-1}; \mu_\theta(\mathbf{x}_t, t, \mathbf{C}), \sigma^2_\theta(t) \mathbf{I}),
    \label{eq:gaussian_reverse_conditional}
\end{equation}
and for discrete models Eq.~\eqref{eq:discrete_reverse} becomes:
\begin{equation}
    p_{\theta}(\mathbf{x}_{t-1}|\mathbf{x}_t, \mathbf{C}) = \text{Cat}(\mathbf{x}_{t-1}; \mathbf{p} = f_\theta(\mathbf{x}_t, t, \mathbf{C})).
    \label{eq:discrete_reverse_conditional}
\end{equation}

\input{figs/architec}
\input{tables/table_diffusion_choice}
\input{tables/table_guidance_choice}
\section{Key Design Decisions}
\label{sec:designs}
\subsection{Denoiser Network Architecture}
We use a U-Net variant~\cite{ronneberger2015u} as the denoiser network, trained to predict the clean mask \(x_0\) rather than directly predicting \(x_{t-1}\), following the \(x_0\) parameterization approach. A point cloud segmentation network~\cite{zhu2021cylindrical} is adapted for denoising, with modifications to handle occupancy data and time embeddings. The time embeddings are implemented using sinusoidal positional encodings, which are further processed by a small neural network consisting of two linear layers and SiLU activation~\cite{elfwing2018sigmoid} to enhance representational capacity.
\subsection{Visual Encoder}
Since directly generating 3D occupancy from 2D images may lead to hallucinations, we employ a BEV (Bird’s-Eye-View) model as the visual encoder to lift 2D images to 3D features. The BEV model runs once during both training and inference. In our experiments we use BEVFormer~\cite{li2022bevformer} as the visual encoder.
\subsection{Options for Diffusion Modeling across Representations}
Given the various representations available for modeling 3D occupancy grid data, it is essential to explore which representation can be more effectively modeled by diffusion models. In this work, we examine three types of representations for diffusion modeling: spatial latent, triplane, and discrete categorical variables, and compare their performance. Fig.~\ref{fig:archi} shows an overview.
\begin{itemize}
    \item \textbf{Spatial Latent.} We encode 3D occupancy grid data into spatial latent representations to reduce computational cost and achieve compactness. The autoencoder consists of a 3D convolutional encoder with skip connections and an implicit MLP decoder with fully-connected layers. After encoding, the diffusion process is applied to the latent representations, as described in Eq.~\eqref{eq:gaussian_forward} and Eq.~\eqref{eq:gaussian_reverse_conditional}. The spatial latent is then decoded to predict semantic class probabilities for reconstructing the occupancy grids.
    \item \textbf{Triplane.} The triplane representation consists of three planes: \( h_{xy} \in \mathbb{R}^{C_h \times X_h \times Y_h} \), \( h_{xz} \in \mathbb{R}^{C_h \times X_h \times Z_h} \), and \( h_{yz} \in \mathbb{R}^{C_h \times Y_h \times Z_h} \), where \(C_h\) is the feature dimension, and \(X_h\), \(Y_h\), and \(Z_h\) represent spatial dimensions. The same encoder as described in the spatial latent section is used, and spatial latent representations are transformed into triplane via average pooling. For a 3D coordinate \(p = (x, y, z)\), the triplane feature \(h(p)\) is the sum of bilinear interpolations from each plane. The diffusion models aim to model these triplane representations, using the process defined in Eq.~\eqref{eq:gaussian_forward} and Eq.~\eqref{eq:gaussian_reverse_conditional}.
    \item \textbf{Discrete categorical variables.} Since occupancy grid data is discrete and categorical, we use a discrete diffusion process for modeling, as defined in Eq.~\eqref{eq:discrete_forward} and Eq.~\eqref{eq:discrete_reverse_conditional}. We add noise to the occupancy grids using uniform transition matrices \(\mathbf{Q_t}\) in Equation Eq.~\eqref{eq:discrete_forward}, as introduced by Nichol and Dhariwal~\cite{sohl2015deep} and adapted for categorical data by Hoogeboom et al.~\cite{hoogeboom2021argmax}. A learnable embedding layer projects the corrupted discrete labels into a high-dimensional continuous feature space for input to the denoiser.
\end{itemize}

We train and validate all three models on the occ3d-NuScenes dataset~\cite{occ3d} with consistent settings. Results in Tab.~\ref{tab:diffusion_targets} indicate that discrete categorical variables outperform the other representations in generative pipeline, as measured by mIoU. This is likely due to the discrete nature of occupancy grid data and potential information loss from the autoencoder. Thus, we use discrete categorical variables to represent occupancy grids for the remainder of this paper.

\input{tables/table_condition_choice}

\subsection{Guidance techniques}
Classifier-free guidance (CFG)~\cite{ho2022classifier} and classifier guidance (CG)~\cite{dhariwal2021diffusion} are commonly used in conditional image generation to enhance the influence of conditions, such as text prompts. CFG adjusts the logits of the conditional (\(\ell_c\)) and unconditional (\(\ell_u\)) models with a guidance scale \(s\), formulated as \(\ell = (s + 1)\ell_c - s\ell_u\). In contrast, CG uses a classifier to compute gradients of the condition log-probability, directing the generation towards the desired outcome. We compare these guidance techniques in our conditional sampling process and find that CFG outperformed CG, as shown in Tab.~\ref{tab:guidance_comparison}. This is likely because the classifier used in CG, a discriminative occupancy prediction model~\cite{fbocc}, has limited classification performance compared to image classifiers, leading to significant errors. Consequently, we choose CFG as the guidance technique for our sampling process.

\subsection{Condition Options}
To leverage the advantages of conditional diffusion models, we conduct experiments using different conditions provided by the base BEV models to assess their impact on occupancy prediction. We evaluate three types of conditions: (1) predictions from the BEV model (C-PR), where the diffusion model refines these initial predictions; (2) logits from the final classifier layer (C-L); and (3) representations before the final classifier layers (C-R), which allows for end-to-end training of both the visual encoder and the diffusion model. As shown in Tab.~\ref{tab:table_condition_choice}, representations before the final classifier layers achieve the best performance. This suggests that these features offer more informative guidance for the diffusion model’s generative process, further enhanced by end-to-end training. Therefore, we use these representations as conditions for the diffusion model.

\input{tables/table_visible_miou}

\section{Fascinating Properties of Diffusion Models for Occupancy Prediction}
In this section, we provide insights and intuitions on modeling occupancy prediction as a diffusion modeling task and highlight its advantages over other discriminative methods.
\subsection{3D Scene Prior}

\input{figs/qualitative}
\input{figs/multimodal}

\noindent Real-world occupancy data frequently encompass intricate 3D structures and detailed object shapes, including pedestrians, buildings, and vegetation. Diffusion models capture and model these complexities as a 3D scene prior more effectively than discriminative methods. This prior enables joint modeling of semantic relationships among voxels, allowing the generation of highly probable outcomes under conditional guidance while preserving scene consistency and plausibility. Consequently, this leads to more precise occupancy perception, outperforming discriminative models, as demonstrated in Sec.~\ref{sec:Compare with sota}. Our qualitative results in Fig.~\ref{fig:qulitative} illustrate the accuracy and reliability of our occupancy predictions.
Moreover, in regions with incomplete observations or occlusions, the inclusion of a 3D scene prior inherently equips the model to infer missing information, resulting in more comprehensive perception outputs (see Sec.~\ref{sec:invisible}). This enhancement is crucial for effective downstream planning, as demonstrated in our experiments in Sec.~\ref{sec:planning}.
\subsection{Robustness to Noisy Data}
During 3D occupancy data annotation, issues such as insufficient observations and sensor noise often lead to imperfect labels, posing challenges for discriminative models and leading to blurred or incomplete predictions. In contrast, diffusion models inherently handle such noise due to their denoising capabilities. Their forward diffusion process acts as an augmentation technique, helping to counteract the impact of noisy occupancy labels. As shown in Sec.~\ref{sec:robustness}, our quantitative analysis demonstrates that diffusion models exhibit superior robustness to noise compared to traditional methods, leading to more accurate occupancy predictions.
\input{figs/more_than_gt}
\vspace{-0.5em}
\subsection{Multi-Modal Occupancy Distributions}
\vspace{-0.3em}
Predicting occupancy grids from multi-view images is inherently ill-posed because there are multiple occupancy configurations that can match the same image observations, resulting in a multi-modal conditional distribution \(q(\mathbf{x}|\mathbf{\mathbf{C}})\). Discriminative models, however, are limited to producing a single prediction and fail to capture this multi-modality, which can hinder downstream tasks such as planning where multiple scenarios need to be considered. In contrast, diffusion models excel at representing multi-modal distributions, allowing them to generate diverse and realistic samples that align with camera observations. As demonstrated in Fig.~\ref{fig:multimodal}, diffusion models effectively capture this variability, providing richer and more accurate predictions.

\vspace{-0.5em}
\subsection{Dynamic Inference Steps}
\vspace{-0.2em}
Leveraging the multi-step sampling process, our approach facilitates dynamic inference, offering a versatile balance between computational resources and the quality of predictions. More discussions are in Sec.~\ref{sec:infer_steps}.

\vspace{-0.3em}
\section{Evaluation}
\vspace{-0.2em}

\subsection{Experimental Setup}
\vspace{-0.2em}
\label{sec:Experimental Setup}
Our experimental setup is based on our best practices outlined in Sec.~\ref{sec:designs}. \textbf{Benchmark.} We evaluate our model on the Occ3D-nuScenes dataset~\cite{occ3d}. This dataset covers a spatial range from -40m to 40m in the X and Y axes, and from -1m to 5.4m in the Z axis, with occupancy labels provided in 0.4m voxel grids across 17 categories. The data collection vehicle is equipped with a LiDAR, five radars, and six cameras, ensuring 360-degree environmental perception.
\input{tables/table_invisible_miou}
\textbf{Settings.} Our framework is designed to be plug-and-play, and given the exceptional performance and generalizability of various mainstream occupancy prediction methods, we selected several off-the-shelf BEV models pretrained using established methodologies as our base models for generating conditions. During training, we used 1000 steps and aligned the remaining training details with those of leading occupancy prediction methods to ensure fairness. For inference, we used a 10-step process and set the guidance scale for CFG to 3.5, unless stated otherwise.
\vspace{-0.7em}
\subsection{Comparison with State-of-the-art Methods}
\vspace{-0.2em}
We evaluated our model using BEVFormer~\cite{li2022bevformer} and PanoOcc~\cite{wang2023panoocc} as visual encoders to produce high-dimensional representations for the generative process. We compared its performance with several popular methods~\cite{monoscene,renderocc,fbocc,huang2021bevdet}. As shown in Tab.~\ref{table:sota_occ_eval}, our approach achieved a 7.05 mIoU improvement over BEVFormer and a 0.97 mIoU gain over PanoOcc. These results underscore the effectiveness and versatility of our generative modeling.
\label{sec:Compare with sota}
\input{tables/table_distant_miou}

\input{tables/table_noisy_miou}

\input{tables/table_planning}
\input{tables/table_steps}

\subsection{Reasoning with Prior in Camera-Invisible Regions}
\label{sec:invisible}
To demonstrate the superior performance of generative models in complex perception tasks like occupancy prediction, we evaluated the mIoU in camera-invisible regions. We defined these regions based on the camera-invisible labels in the Occ3D-nuScenes datasets and compared our method against state-of-the-art approaches, as detailed in Tab.~\ref{tab:camera_invisble}.

Real-world occupancy datasets often rely on annotations derived from aggregated LiDAR point clouds, which may not cover all invisible regions, limiting the evaluation across entire scenes. However, qualitative results indicate that our method consistently delivers realistic and reasonable predictions throughout the entire scene, as illustrated in Fig.~\ref{fig:more_than_gt}.

\subsection{Performance In Noisy Regions}
\label{sec:robustness}
To evaluate the model’s performance in noisy regions, we evaluated the mIoU in noise-prone distant areas (20 meters away), as detailed in Tab.~\ref{tab:distant}. We also calculated the visibility probabilities of all voxels near the ego vehicle across the Occ3D-nuScenes~\cite{occ3d} dataset’s training set and evaluated mIoU in regions with lower visibility probabilities, as shown in Tab.~\ref{tab:noisy prob}. Our results show that generative models outperform discriminative methods in these regions, highlighting their superior ability to learn the true scene distribution and maintain robustness in high-noise environments.

\subsection{Inference Steps.}
\label{sec:infer_steps}
In our experiments, we found that maintaining the complete sequence of inference steps identical to the training phase significantly reduces performance. This issue may be due to discrepancies between the training and testing distributions~\cite{lai2023ddps}. Notably, using only the initial few steps and leveraging the reconstructed $x_0$ led to performance saturation. We observed that performance peaks around 10 to 15 steps. The trade-off between performance and the number of inference steps for DiffOcc is illustrated in Tab.~\ref{tab:infer_steps}.
\vspace{-0.4em}
\subsection{planning}
\vspace{-0.2em}
\label{sec:planning}
We evaluate the quality of occupancy prediction from a new perspective: its impact on downstream planning tasks. The ultimate goal of perception modules in autonomous driving is to support planning. However, the commonly used IoU metric, calculated only on visible grids (filtered by a visible mask), overlooks the importance of predicting a complete, physically consistent, and realistic occupancy scene—an essential factor for decision-making in planning. We argue that our diffusion-based method provides more informative occupancy predictions for planning modules by leveraging implicit 3D scene priors.
To validate this, we modify a simple planning module based on UniAD~\cite{hu2023_uniad}, replacing its Bird's-Eye-View (BEV) features with ground-truth occupancy annotations. During evaluation, we assess the effectiveness of occupancy predictions from different models. As shown in Tab.~\ref{tab:sota-plan}, our method outperforms the discriminative model both with and without visible masks. When trained and tested without visible masks, the performance of the discriminative model drops significantly, whereas our method surpasses even the ground-truth occupancy annotations. This demonstrates that our model offers more informative and comprehensive environment perception.

\vspace{-0.2em}
\section{Related Works}
\vspace{-0.2em}
\subsection{3D Occupancy Prediction and Completion}
With the growing importance of vision-centric autonomous driving systems, an increasing number of researchers are focusing on 3D occupancy prediction tasks~\cite{occ3d, openoccupancy, occformer, monoscene, surroundocc, pointocc, renderocc, simpleocc, scenerf, tpvformer, fbocc, voxformer, sceneasocc, mapprior}. A related task is Semantic Scene Completion (SSC), which aims to estimate a dense semantic space from partial observations~\cite{armeni2017joint, dai2017scannet, liao2022kitti, lee2023diffusion}. Although both tasks produce similar outputs, SSC emphasizes reconstructing 3D scene geometry and semantics from sparse data, whereas 3D occupancy prediction focuses on accurately representing the occupancy of 3D space, particularly for both dynamic and static objects within the sensor-visible range. Our model utilizes generative approaches to address prediction tasks and can also be applied to completion tasks.
\subsection{Generative Models for Autonomous Driving and Robotics}
Generative models have found extensive applications in autonomous driving and robotics~\cite{bevcontrol, bevgen, drivingdiffusion, magicdrive, zhang2023learning, wen2023panacea, xiong2023learning,lange2022lopr}. Specifically, in the realm of perception, MapPrior~\cite{mapprior} introduced a novel BEV perception framework that integrates a traditional discriminative BEV perception model with a learned generative model for semantic map layouts. UltraLiDAR~\cite{xiong2023learning} pioneered the use of VQ-VAE to complete and generate realistic LiDAR point clouds. Copilot4D~\cite{zhang2023learning} developed a discrete-diffusion-based model tailored for 4D LiDAR point clouds, achieving state-of-the-art performance. Similarly, DiffBEV~\cite{zou2023diffbev} leveraged diffusion models to generate a more comprehensive BEV representation. The most relevant work to ours is OccGen\cite{wang2024occgen}, but it treats diffusion models as a coarse-to-fine process while overlooking many properties of diffusion models in occupancy prediction.
\vspace{-0.2em}
\section{Conclusion \& Discussion}
\vspace{-0.2em}
\noindent\textbf{Conclusion.} 
Our experiments demonstrate the superior performance of Diffocc in challenging scenarios, offering more accurate and realistic predictions. This advance not only enhances perception capabilities but also benefits downstream planning tasks, highlighting the potential of generative modeling for improving autonomous systems.

\noindent\textbf{Disccusion.} \textit{Inference Latency} is also an important consideration. Tab.~\ref{tab:infer_steps} shows that our model can achieve good performance with just 1-2 sampling steps. Utilizing a faster base model and acceleration techniques for diffusion models can further enhance the applicability of generative models for occupancy prediction. \textit{Hallucination} is a common concern for generative models; however, the mIoU, as a discriminative metric, shows that generative modeling achieves superior perception accuracy compared to discriminative models. Moreover, the performance improvements in planning tasks further demonstrate that this modeling approach does not induce hallucinations detrimental to downstream tasks.




\section*{ACKNOWLEDGMENT}
This work is supported by National Key R\&D Program of China (2022ZD0161700) and Tsinghua University Initiative Scientific Research Program.




\clearpage
\bibliographystyle{IEEEtran}
\bibliography{root}

\end{document}

%% file: figs/teaser.tex
\begin{figure*}[t]
  \centering
    \includegraphics[width=\textwidth]{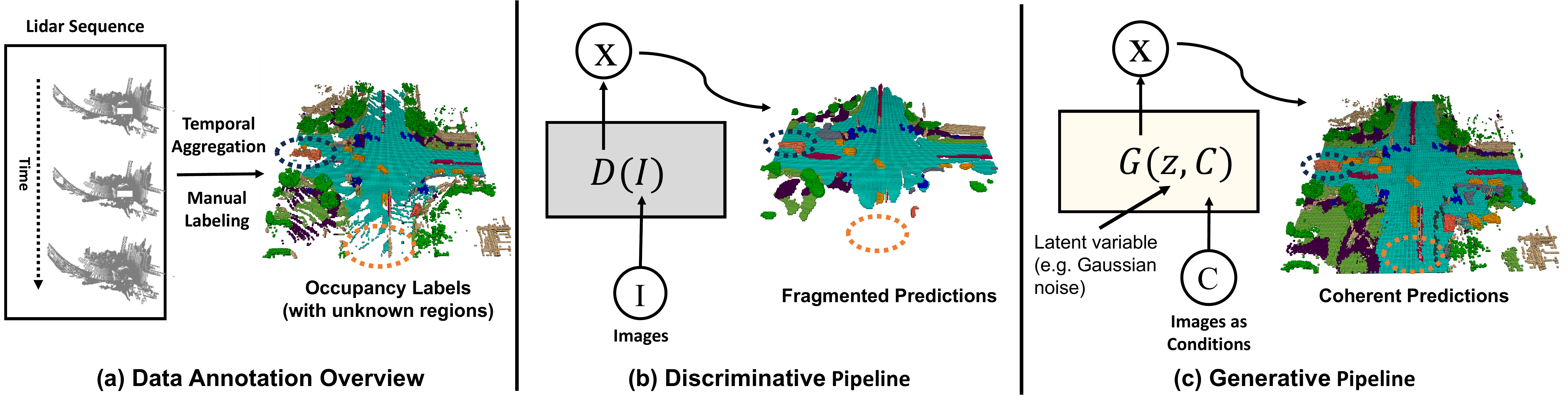}
    \caption{The diagram illustrates the occupancy data production process (a), the discriminative pipeline (b), and the generative pipeline (c). Black and orange circles highlight the deficiencies in the results from both the data production process and the discriminative pipeline, in contrast to the more reasonable results produced by the generative pipeline.}
\label{fig:teaser}
\vspace{-15pt}
\end{figure*}

%% file: figs/architec.tex
\begin{figure*}[t]
  \centering
    \includegraphics[width=\textwidth]{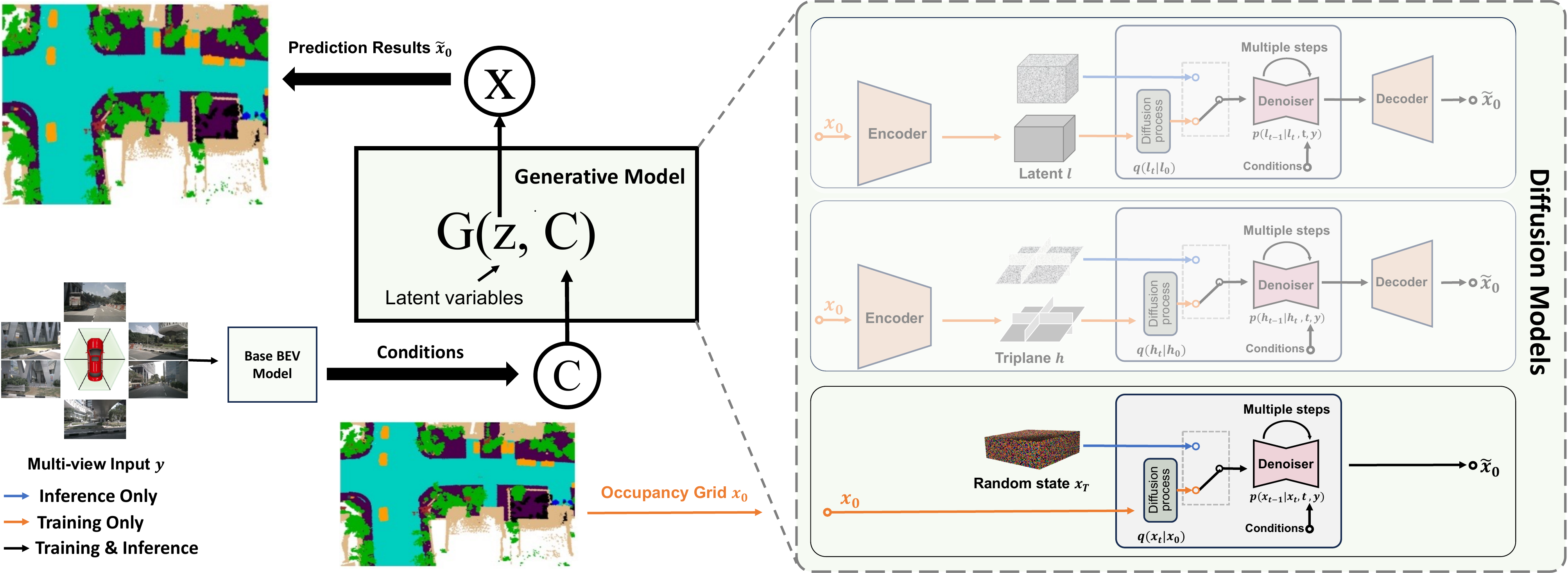}
    \caption{An overview of using diffusion models for occupancy prediction. A base BEV model is employed to encode the input into high-dimensional features, which serve as conditions for the diffusion models. The diffusion model is then able to model different representations of the occupancy grid data.}
\label{fig:archi}
\vspace{-2em}
\end{figure*}

%% file: tables/table_diffusion_choice.tex
\begin{table}[t]
  \centering
  \setlength{\tabcolsep}{3pt} 
  \caption{Comparison of different occupancy representations for modeling with diffusion models. “DiffOcc(***)” denoting the adaptation of diffusion models to the specified representations denoted as “***”.}
  \resizebox{\columnwidth}{!}{
    \begin{tabular}{c|cccc}
    \toprule
      Model& BEVFormer & DiffOcc(latent) & DiffOcc(triplane) & DiffOcc(discrete) \\
      \hline
      \textbf{mIoU} & 23.67 & 30.65 & 30.23 & \textbf{31.78} \\
    \bottomrule
    \end{tabular}
  }
  
  \label{tab:diffusion_targets}
    \vspace{-5pt}

\end{table}

%% file: tables/table_guidance_choice.tex
\begin{table}[t]
  \centering
  \caption{Comparison of mIoU for different guidance techniques and scales, obtained using discrete diffusion models.}
  \label{tab:guidance_comparison}
  \begin{tabular}{l|c|c|c|c}
    \toprule
    Guidance Tech & Scale 0.5 & Scale 1 & Scale 2 & Scale 3.5 \\
    \midrule
    CFG & 27.34 & 31.78 & 31.98 &\textbf{32.45} \\
    CG  & 26.50 & 29.45 & 29.98 & 29.97 \\
    \bottomrule
  \end{tabular}
    \vspace{-15pt}
\end{table}

%% file: tables/table_condition_choice.tex
\begin{table}[t]
\centering
\caption{Comparison of mIoU for different conditions. Results obtained using conditional sampling with a CFG scale of 3.5.}
  \vspace{-10pt}

  \begin{tabular}{c|ccc}
    \toprule
     Cond. Option & C-PR & C-L & C-R \\
    \midrule
    \textbf{mIoU} & 32.45 & 31.77 & \textbf{40.63} \\
    \bottomrule
  \end{tabular}

\label{tab:table_condition_choice}
  \vspace{-15pt}

\end{table}

%% file: tables/table_visible_miou.tex
\vspace{15pt}
\begin{table*}[h]
\footnotesize
\small
\setlength{\tabcolsep}{0.005\linewidth}
\caption{\textbf{3D Occupancy Prediction Performance on the Occ3D-nuScenes Validation Dataset.} The evaluation is conducted using a LiDAR mask. The symbol~$\bullet$ means the backbone is pretrained using the nuScense segmentation.
“Cons. Veh” represents construction vehicle, and “Driv. Sur” is short for driveable surface.~$\ast$ indicates our own re-implementation.
“DiffOcc (x)” denotes the use of representations obtained from “x” as conditions for diffusion models.}
\def\mystrut{\rule{0pt}{1.5\normalbaselineskip}}
\centering
\begin{adjustbox}{width=\textwidth,center}
\begin{tabular}{l | c c | >{\columncolor{gray!20}}c | r r r r r r r r r r r r r r r r r r}
    \toprule[1.5pt]
    Method 
    & \rotatebox{90}{Backbone}
    & \rotatebox{90}{Image size}
    & \rotatebox{90}{\textbf{mIoU}}$\uparrow$ 
    & \rotatebox{90}{Others}$\uparrow$ 
    & \rotatebox{90}{Barrier}$\uparrow$
    & \rotatebox{90}{Bicycle}$\uparrow$ 
    & \rotatebox{90}{Bus}$\uparrow$ 
    & \rotatebox{90}{Car}$\uparrow$ 
    & \rotatebox{90}{Cons. Veh}$\uparrow$ 
    & \rotatebox{90}{Motorcycle}$\uparrow$ 
    & \rotatebox{90}{Pedestrian}$\uparrow$ 
    & \rotatebox{90}{Traffic cone}$\uparrow$ 
    & \rotatebox{90}{Trailer}$\uparrow$ 
    & \rotatebox{90}{Truck}$\uparrow$ 
    & \rotatebox{90}{Driv. Sur}$\uparrow$ 
    & \rotatebox{90}{Other flat}$\uparrow$ 
    & \rotatebox{90}{Sidewalk}$\uparrow$ 
    & \rotatebox{90}{Terrain}$\uparrow$ 
    & \rotatebox{90}{Manmade}$\uparrow$ 
    & \rotatebox{90}{Vegetation}$\uparrow$ 
    \\
    \midrule
    MonoScene~\cite{monoscene} & R101 & 900$\times$1600 & 6.0 & 1.7 & 7.2 & 4.2 & 4.9 & 9.3 & 5.6 & 3.9 & 3.0 & 5.9 & 4.4 & 7.1 & 14.9 & 6.3 & 7.9 & 7.4 & 1.0 & 7.6 \\
    BEVFormer~\cite{li2022bevformer} & R101 & 900$\times$1600&22.36 & 2.7 & 35.57 & 8.43 & 32.09 & 40.19 & 12.15 & 14.99 & 17.04 & 16.18 & 16.49 & 27.88 & 48.75 & 27.64 & 28.87 & 25.15 & 11.16 & 14.84  \\
    RenderOcc~\cite{renderocc} & SwinB & 512$\times$1408 & 24.46 & 4.74 & 26.64 & 12.48 & 23.13 & 22.25 & 13.29 & 10.97 & 13.92 & 13.18 & 23.28 & 21.21 & 67.62 & 33.02 & 43.37 & 44.18 & 19.11 & 23.42 \\
    FB-Occ~\cite{fbocc} & R101 & 256$\times$704 & 38.33 & 8.52 & 43.85 & 26.25 & 42.79 & 49.59 & 22.94 & 27.65 & 27.73 & 26.98 & 34.41 & 36.73 & 78.36 & 41.41 & 47.01 & 49.80 & 44.89 & 42.54 \\
    BEVDet~\cite{huang2021bevdet} & SwinB & 512$\times$1408 & 39.08 & 7.17 & 46.11 & 22.09 & 46.2 & 51.57 & 24.08 & 25.15 & 27.11 & 25.1 & 34.77 & 37.89 & 78.92 & 42.13 & 50.63 & 50.54 & 49.04 & 45.91 \\
    PanoOcc~\cite{wang2023panoocc} & R101$\bullet$ & 864$\times$1600 & 42.32 & 8.37 & 49.86 & 28.11 & 51.45 & 58.97 & 24.82 & 34.6 & 32.29 & 30.15 & 37.75 & 46.77 & 80.8 & 42.97 & 51.25 & 45.38 & 44.33 & 42.32 \\
    \midrule
    BEVFormer$\ast$~\cite{li2022bevformer} & R101 & 900$\times$1600 & 35.62 & 4.96 & 41.48 & 15.81 & 44.65 & 50.97 & 20.9 & 20.74 & 25.87 & 22.32 & 33.11 & 35.59 & 78.05 & 39.04 & 44.49 & 48.17&41.08&  38.25 \\
    PanoOcc~\cite{wang2023panoocc} & R101 & 864$\times$1600 & 42.15 & 8.51 & 49.22 & 28.35 & 50.37 & 58.48 & 25.58 & 34.3 & 31.16 & 30.49 & 38.38 & 45.37 & 81.2 & 44.07 & 50.8 & 51.63 & 44.83 & 42.15 \\
    \midrule
    \bf{DiffOcc (BEVFormer$\ast$)} & R101                    & 900$\times$1600 & \textbf{42.67} & 8.59 & 47.61 & 26.02 & 54.21 & 58.11 & 26.2 & 29.55 & 31.1 & 29.61 & 38.93 & 46.19 & 81.23 & 44.01 & 50.99 & 51.6 & 51.93 & 49.56 \\
    \bf{DiffOcc (PanoOcc)}    & R101 & 864$\times$1600 & \textbf{43.08} & 7.86 & 48.67 & 25.66 & 55.73 & 57.14 & 27.17 & 32.33 & 30.06 & 30.1 & 40.42 & 48.48 & 82.01 & 43.98 & 52.06 & 52.03 & 49.98 & 48.69 \\

\bottomrule[1.5pt]
\end{tabular}
\end{adjustbox}
\vspace{0.1cm}
\label{table:sota_occ_eval}
\vspace{-15pt}
\end{table*}

%% file: figs/qualitative.tex
\begin{figure}[t]
  \centering
    \includegraphics[width=1.0\linewidth]{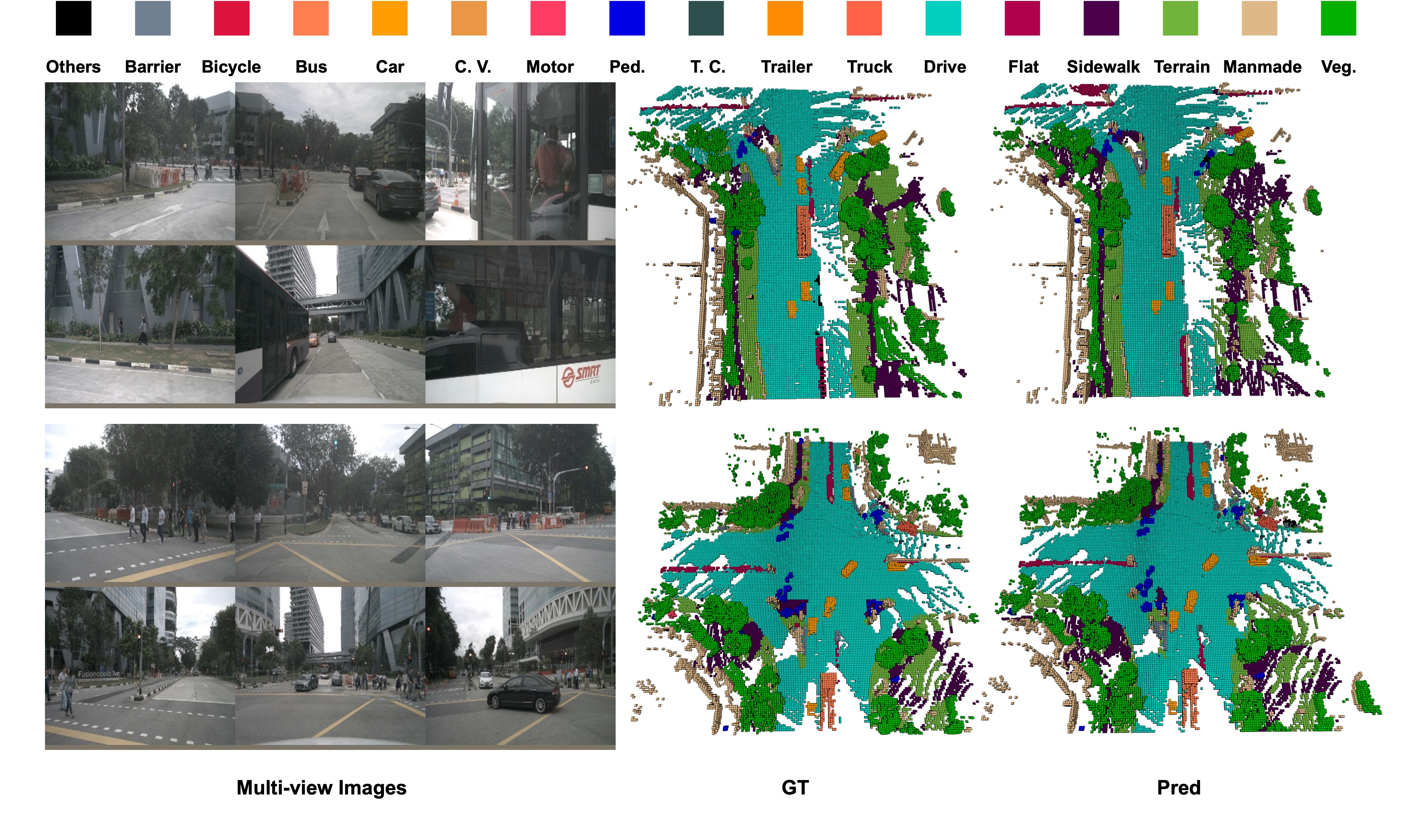}
     \vspace{-20pt}
    \caption{\textbf{Qualitative results on Occ3D-nuScenes validation set}. }
  \label{fig:qulitative}
  \vspace{-15pt}
\end{figure}

%% file: figs/multimodal.tex
\begin{figure}[t]
  \centering
    \includegraphics[width=1.0\linewidth]{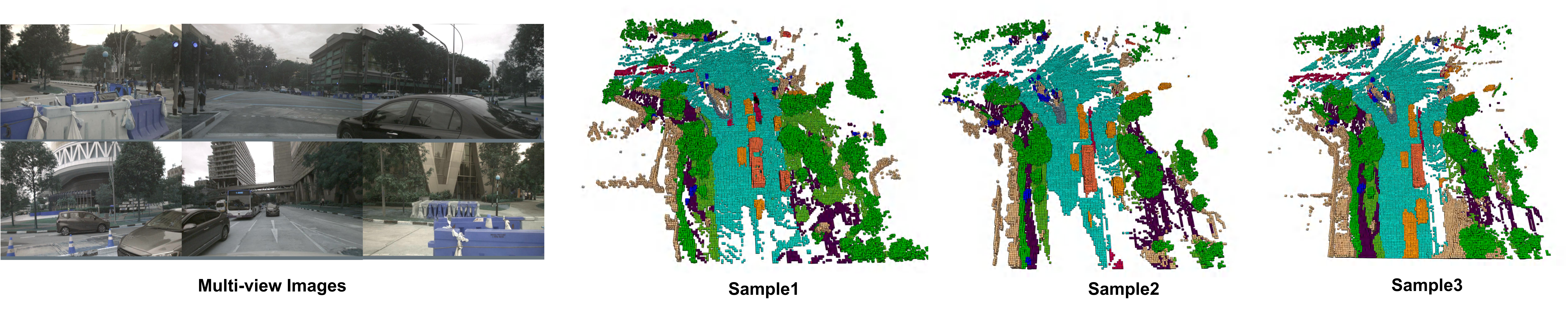}
    \vspace{-15pt}
    \caption{\textbf{Qualitative results of multi-modality.}}
  \label{fig:multimodal}
  \vspace{-20pt}
\end{figure}

%% file: figs/more_than_gt.tex
\begin{figure}[t]
  \centering
    \includegraphics[width=1.0\linewidth]{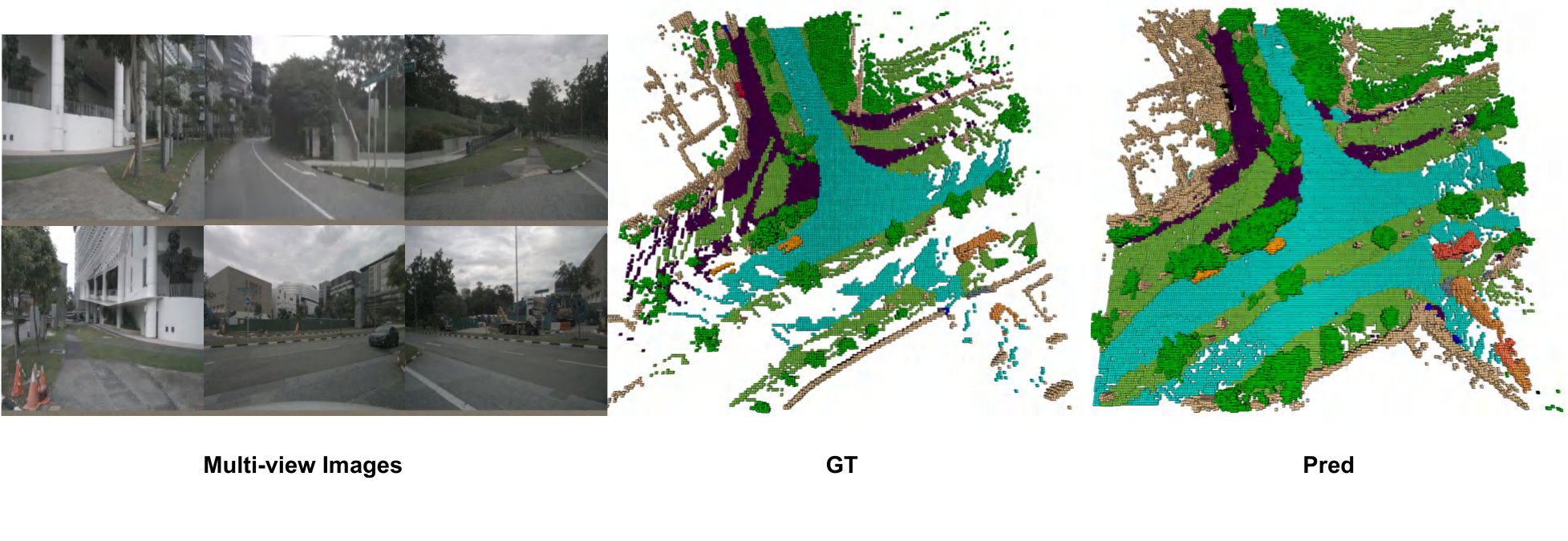}
      \vspace{-20pt}
    \caption{\textbf{Ground Truth vs. Predictions}, Our model provides denser and more coherent occupancy estimations compared to point cloud-derived ground truths (for instance, it includes complete drivable surfaces).}
  \label{fig:more_than_gt}
  \vspace{-15pt}
\end{figure}

%% file: tables/table_invisible_miou.tex
\begin{table}[t]
\centering
\caption{3D occupancy prediction performance in camera-invisible regions.}

\begin{tabular}{l | c  | c }
    \toprule[1.5pt]
    Method 
    & \rotatebox{90}{Backbone}
    & \rotatebox{90}{\textbf{mIoU}} 
    \\
    \midrule
    BEVFormer$\ast$~\cite{li2022bevformer} & R101 & 21.42  \\
    BEVDet~\cite{huang2021bevdet} & SwinB  & 21.79 \\
    FB-Occ~\cite{fbocc} & R101  & 24.84 \\
    RenderOcc~\cite{renderocc} & SwinB  & 22.57 \\
    PanoOcc~\cite{wang2023panoocc} & R101 & 30.13 \\
    \midrule

    \bf{DiffOcc (BEVFormer$\ast$)} & R101  &  \textbf{36.45} \\
    \bf{DiffOcc (PanoOcc)}    & R101  &  \textbf{36.83}\\

\bottomrule[1.5pt]
\end{tabular}
\label{tab:camera_invisble}
  \vspace{-12pt}
\end{table}

%% file: tables/table_distant_miou.tex
\begin{table}[t]
\centering
\caption{3D occupancy prediction performance at long distances.}
\begin{tabular}{l | c | c }
    \toprule[1.5pt]
    Method 
    & Backbone
    & \textbf{mIoU}  
    \\
    \midrule
    BEVFormer$\ast$ & R101  & 26.03  \\
    PanoOcc & R101 &  30.13 \\
    \midrule
    \bf{DiffOcc (BEVFormer$\ast$)} & R101 &  \textbf{30.73} \\
    \bf{DiffOcc (PanoOcc)}    & R101 &   \textbf{32.04}\\

\bottomrule[1.5pt]
\end{tabular}

\label{tab:distant}
\vspace{-5pt}
\end{table}

%% file: tables/table_noisy_miou.tex
\begin{table}[t]
\centering
\caption{mIoU scores of different methods across various visible probability values. Each column represents the mIoU scores for voxels where the visibility probability is below the specified threshold, showcasing the performance of each method under varying degrees of visibility constraints.}
\begin{tabular}{lcccc}
\toprule
Method & 5\%  & 10\%  & 20\%   & 50\%   \\
\midrule
 BEVFormer$\ast$& 28.1 &  29.55& 32.26 & 33.59 \\
\textbf{DiffOcc (BEVFormer$\ast$)}& \textbf{33.82} & \textbf{36.02} & \textbf{39.69} & \textbf{41.64} \\
\bottomrule
\end{tabular}

\label{tab:noisy prob}
\vspace{-20pt}
\end{table}

%% file: tables/table_planning.tex
\begin{table*}[t] 
\vspace{1pt}
\centering
\caption{Comparison of occupancy prediction effectiveness across different models. "G.T. Occ." refers to the use of ground-truth occupancy annotations. "BEVFormer" represents results from the standard BEVFormer model, while "DiffOcc" uses our proposed diffusion-based objectives. $^\dagger$ indicates that no visible masks were applied during training or evaluation.}

\begin{tabular}{l|ccc>{\columncolor{gray!20}}c|ccc>{\columncolor{gray!20}}c}
\toprule
\multirow{2}{*}{Method} & 
\multicolumn{4}{c|}{\textbf{L2 (m)} $\downarrow$} & 
\multicolumn{4}{c}{\textbf{Collision} (\%) $\downarrow$} \\
& 1s & 2s & 3s & Avg. & 1s & 2s & 3s & Avg. \\
\midrule
G.T. Occ. & 1.62 & 3.02 & 4.20 & 2.95 & 2.97 & 3.77 & 4.53 & 3.76 \\
BEVFormer & 1.82 & 3.39 & 4.73 & 3.31 & 2.52 & 3.26 & 4.72 & 3.50 \\
\textbf{DiffOcc (BEVFormer)} & 1.66 & 3.08 & 4.29 & 3.01 & 2.39 & 3.26 & 4.37 & 3.34 \\
BEVFormer$^\dagger$ & 2.46 & 5.27 & 7.71 & 5.15 & \textbf{1.66} & 4.29 & 8.22 & 4.72 \\
\textbf{DiffOcc (BEVFormer)$^\dagger$} & \textbf{1.56} & \textbf{2.91} & \textbf{4.12} & \textbf{2.87} & 1.86 & \textbf{2.74} & \textbf{4.87} & \textbf{3.16} \\
\bottomrule
\end{tabular}
\vspace{-15pt}
\label{tab:sota-plan}
\end{table*}

%% file: tables/table_steps.tex
\begin{table}[t]
\centering
\caption{mIoU scores of different sample steps $t$ during inference.}

  \begin{tabular}{c|cccccc}
    \toprule
    Steps & 1 & 2 & 10 & \textbf{15} & 20 & 50 \\
    \midrule
    \textbf{mIoU} & 40.48 & 40.63 & 42.12 & \textbf{42.14} & 41.99 & 40.47 \\
    \bottomrule
  \end{tabular}

\label{tab:infer_steps}
\vspace{-15pt}
\end{table}

%% file: root.bbl
\begin{thebibliography}{10}
\providecommand{\url}[1]{#1}
\csname url@rmstyle\endcsname
\providecommand{\newblock}{\relax}
\providecommand{\bibinfo}[2]{#2}
\providecommand\BIBentrySTDinterwordspacing{\spaceskip=0pt\relax}
\providecommand\BIBentryALTinterwordstretchfactor{4}
\providecommand\BIBentryALTinterwordspacing{\spaceskip=\fontdimen2\font plus
\BIBentryALTinterwordstretchfactor\fontdimen3\font minus \fontdimen4\font\relax}
\providecommand\BIBforeignlanguage[2]{{%
\expandafter\ifx\csname l@#1\endcsname\relax
\typeout{** WARNING: IEEEtran.bst: No hyphenation pattern has been}%
\typeout{** loaded for the language `#1'. Using the pattern for}%
\typeout{** the default language instead.}%
\else
\language=\csname l@#1\endcsname
\fi
#2}}

\bibitem{liao2022kitti}
Y.~Liao, J.~Xie, and A.~Geiger, ``Kitti-360: A novel dataset and benchmarks for urban scene understanding in 2d and 3d,'' \emph{IEEE Transactions on Pattern Analysis and Machine Intelligence}, 2022.

\bibitem{semantickitti}
J.~Behley, M.~Garbade, A.~Milioto, J.~Quenzel, S.~Behnke, C.~Stachniss, and J.~Gall, ``Semantickitti: A dataset for semantic scene understanding of lidar sequences,'' in \emph{Proceedings of the IEEE/CVF international conference on computer vision}, 2019, pp. 9297--9307.

\bibitem{occ3d}
X.~Tian, T.~Jiang, L.~Yun, Y.~Wang, Y.~Wang, and H.~Zhao, ``Occ3d: A large-scale 3d occupancy prediction benchmark for autonomous driving,'' \emph{arXiv preprint arXiv:2304.14365}, 2023.

\bibitem{ddpm}
J.~Ho, A.~Jain, and P.~Abbeel, ``Denoising diffusion probabilistic models,'' \emph{Advances in neural information processing systems}, vol.~33, pp. 6840--6851, 2020.

\bibitem{ddim}
J.~Song, C.~Meng, and S.~Ermon, ``Denoising diffusion implicit models,'' \emph{arXiv preprint arXiv:2010.02502}, 2020.

\bibitem{hoogeboom2021argmax}
E.~Hoogeboom, D.~Nielsen, P.~Jaini, P.~Forr{\'e}, and M.~Welling, ``Argmax flows and multinomial diffusion: Learning categorical distributions,'' \emph{Advances in Neural Information Processing Systems}, vol.~34, pp. 12\,454--12\,465, 2021.

\bibitem{austin2021structured}
J.~Austin, D.~D. Johnson, J.~Ho, D.~Tarlow, and R.~Van Den~Berg, ``Structured denoising diffusion models in discrete state-spaces,'' \emph{Advances in Neural Information Processing Systems}, vol.~34, pp. 17\,981--17\,993, 2021.

\bibitem{ronneberger2015u}
O.~Ronneberger, P.~Fischer, and T.~Brox, ``U-net: Convolutional networks for biomedical image segmentation,'' in \emph{Medical image computing and computer-assisted intervention--MICCAI 2015: 18th international conference, Munich, Germany, October 5-9, 2015, proceedings, part III 18}.\hskip 1em plus 0.5em minus 0.4em\relax Springer, 2015, pp. 234--241.

\bibitem{zhu2021cylindrical}
X.~Zhu, H.~Zhou, T.~Wang, F.~Hong, Y.~Ma, W.~Li, H.~Li, and D.~Lin, ``Cylindrical and asymmetrical 3d convolution networks for lidar segmentation,'' in \emph{Proceedings of the IEEE/CVF conference on computer vision and pattern recognition}, 2021, pp. 9939--9948.

\bibitem{elfwing2018sigmoid}
S.~Elfwing, E.~Uchibe, and K.~Doya, ``Sigmoid-weighted linear units for neural network function approximation in reinforcement learning,'' \emph{Neural networks}, vol. 107, pp. 3--11, 2018.

\bibitem{li2022bevformer}
Z.~Li, W.~Wang, H.~Li, E.~Xie, C.~Sima, T.~Lu, Y.~Qiao, and J.~Dai, ``Bevformer: Learning bird’s-eye-view representation from multi-camera images via spatiotemporal transformers,'' in \emph{European conference on computer vision}.\hskip 1em plus 0.5em minus 0.4em\relax Springer, 2022, pp. 1--18.

\bibitem{sohl2015deep}
J.~Sohl-Dickstein, E.~Weiss, N.~Maheswaranathan, and S.~Ganguli, ``Deep unsupervised learning using nonequilibrium thermodynamics,'' in \emph{International conference on machine learning}.\hskip 1em plus 0.5em minus 0.4em\relax PMLR, 2015, pp. 2256--2265.

\bibitem{ho2022classifier}
J.~Ho and T.~Salimans, ``Classifier-free diffusion guidance,'' \emph{arXiv preprint arXiv:2207.12598}, 2022.

\bibitem{dhariwal2021diffusion}
P.~Dhariwal and A.~Nichol, ``Diffusion models beat gans on image synthesis,'' \emph{Advances in neural information processing systems}, vol.~34, pp. 8780--8794, 2021.

\bibitem{fbocc}
Z.~Li, Z.~Yu, D.~Austin, M.~Fang, S.~Lan, J.~Kautz, and J.~M. Alvarez, ``Fb-occ: 3d occupancy prediction based on forward-backward view transformation,'' \emph{arXiv preprint arXiv:2307.01492}, 2023.

\bibitem{monoscene}
A.-Q. Cao and R.~de~Charette, ``Monoscene: Monocular 3d semantic scene completion,'' in \emph{CVPR}, 2022, pp. 3991--4001.

\bibitem{renderocc}
M.~Pan, J.~Liu, R.~Zhang, P.~Huang, X.~Li, L.~Liu, and S.~Zhang, ``Renderocc: Vision-centric 3d occupancy prediction with 2d rendering supervision,'' \emph{arXiv preprint arXiv:2309.09502}, 2023.

\bibitem{huang2021bevdet}
J.~Huang, G.~Huang, Z.~Zhu, Y.~Ye, and D.~Du, ``Bevdet: High-performance multi-camera 3d object detection in bird-eye-view,'' \emph{arXiv preprint arXiv:2112.11790}, 2021.

\bibitem{wang2023panoocc}
Y.~Wang, Y.~Chen, X.~Liao, L.~Fan, and Z.~Zhang, ``Panoocc: Unified occupancy representation for camera-based 3d panoptic segmentation,'' \emph{arXiv preprint arXiv:2306.10013}, 2023.

\bibitem{lai2023ddps}
Z.~Lai, Y.~Duan, J.~Dai, Z.~Li, Y.~Fu, H.~Li, Y.~Qiao, and W.~Wang, ``Denoising diffusion semantic segmentation with mask prior modeling,'' \emph{arXiv preprint arXiv:2306.01721}, 2023.

\bibitem{hu2023_uniad}
Y.~Hu, J.~Yang, L.~Chen, K.~Li, C.~Sima, X.~Zhu, S.~Chai, S.~Du, T.~Lin, W.~Wang, L.~Lu, X.~Jia, Q.~Liu, J.~Dai, Y.~Qiao, and H.~Li, ``Planning-oriented autonomous driving,'' in \emph{Proceedings of the IEEE/CVF Conference on Computer Vision and Pattern Recognition}, 2023.

\bibitem{openoccupancy}
X.~Wang, Z.~Zhu, W.~Xu, Y.~Zhang, Y.~Wei, X.~Chi, Y.~Ye, D.~Du, J.~Lu, and X.~Wang, ``Openoccupancy: A large scale benchmark for surrounding semantic occupancy perception,'' in \emph{ICCV}, 2023.

\bibitem{occformer}
Y.~Zhang, Z.~Zhu, and D.~Du, ``Occformer: Dual-path transformer for vision-based 3d semantic occupancy prediction,'' in \emph{ICCV}, 2023.

\bibitem{surroundocc}
Y.~Wei, L.~Zhao, W.~Zheng, Z.~Zhu, J.~Zhou, and J.~Lu, ``Surroundocc: Multi-camera 3d occupancy prediction for autonomous driving,'' in \emph{ICCV}, 2023, pp. 21\,729--21\,740.

\bibitem{pointocc}
S.~Zuo, W.~Zheng, Y.~Huang, J.~Zhou, and J.~Lu, ``Pointocc: Cylindrical tri-perspective view for point-based 3d semantic occupancy prediction,'' \emph{arXiv preprint arXiv:2308.16896}, 2023.

\bibitem{simpleocc}
W.~Gan, N.~Mo, H.~Xu, and N.~Yokoya, ``A simple attempt for 3d occupancy estimation in autonomous driving,'' \emph{arXiv preprint arXiv:2303.10076}, 2023.

\bibitem{scenerf}
A.-Q. Cao and R.~de~Charette, ``Scenerf: Self-supervised monocular 3d scene reconstruction with radiance fields,'' in \emph{ICCV}, 2023, pp. 9387--9398.

\bibitem{tpvformer}
Y.~Huang, W.~Zheng, Y.~Zhang, J.~Zhou, and J.~Lu, ``Tri-perspective view for vision-based 3d semantic occupancy prediction,'' in \emph{CVPR}, 2023, pp. 9223--9232.

\bibitem{voxformer}
Y.~Li, Z.~Yu, C.~Choy, C.~Xiao, J.~M. Alvarez, S.~Fidler, C.~Feng, and A.~Anandkumar, ``Voxformer: Sparse voxel transformer for camera-based 3d semantic scene completion,'' in \emph{Proceedings of the IEEE/CVF Conference on Computer Vision and Pattern Recognition}, 2023, pp. 9087--9098.

\bibitem{sceneasocc}
W.~Tong, C.~Sima, T.~Wang, L.~Chen, S.~Wu, H.~Deng, Y.~Gu, L.~Lu, P.~Luo, D.~Lin, \emph{et~al.}, ``Scene as occupancy,'' in \emph{ICCV}, 2023, pp. 8406--8415.

\bibitem{mapprior}
X.~Zhu, V.~Zyrianov, Z.~Liu, and S.~Wang, ``Mapprior: Bird's-eye view map layout estimation with generative models,'' in \emph{Proceedings of the IEEE/CVF International Conference on Computer Vision}, 2023, pp. 8228--8239.

\bibitem{armeni2017joint}
I.~Armeni, S.~Sax, A.~R. Zamir, and S.~Savarese, ``Joint 2d-3d-semantic data for indoor scene understanding,'' \emph{arXiv preprint arXiv:1702.01105}, 2017.

\bibitem{dai2017scannet}
A.~Dai, A.~X. Chang, M.~Savva, M.~Halber, T.~Funkhouser, and M.~Nie{\ss}ner, ``Scannet: Richly-annotated 3d reconstructions of indoor scenes,'' in \emph{Proceedings of the IEEE conference on computer vision and pattern recognition}, 2017, pp. 5828--5839.

\bibitem{lee2023diffusion}
J.~Lee, W.~Im, S.~Lee, and S.-E. Yoon, ``Diffusion probabilistic models for scene-scale 3d categorical data,'' \emph{arXiv preprint arXiv:2301.00527}, 2023.

\bibitem{bevcontrol}
K.~Yang, E.~Ma, J.~Peng, Q.~Guo, D.~Lin, and K.~Yu, ``Bevcontrol: Accurately controlling street-view elements with multi-perspective consistency via bev sketch layout,'' \emph{arXiv preprint arXiv:2308.01661}, 2023.

\bibitem{bevgen}
A.~Swerdlow, R.~Xu, and B.~Zhou, ``Street-view image generation from a bird's-eye view layout,'' \emph{IEEE Robotics and Automation Letters}, 2024.

\bibitem{drivingdiffusion}
X.~Li, Y.~Zhang, and X.~Ye, ``Drivingdiffusion: Layout-guided multi-view driving scene video generation with latent diffusion model,'' \emph{arXiv preprint arXiv:2310.07771}, 2023.

\bibitem{magicdrive}
R.~Gao, K.~Chen, E.~Xie, L.~Hong, Z.~Li, D.-Y. Yeung, and Q.~Xu, ``Magicdrive: Street view generation with diverse 3d geometry control,'' \emph{arXiv preprint arXiv:2310.02601}, 2023.

\bibitem{zhang2023learning}
L.~Zhang, Y.~Xiong, Z.~Yang, S.~Casas, R.~Hu, and R.~Urtasun, ``Learning unsupervised world models for autonomous driving via discrete diffusion,'' \emph{arXiv preprint arXiv:2311.01017}, 2023.

\bibitem{wen2023panacea}
Y.~Wen, Y.~Zhao, Y.~Liu, F.~Jia, Y.~Wang, C.~Luo, C.~Zhang, T.~Wang, X.~Sun, and X.~Zhang, ``Panacea: Panoramic and controllable video generation for autonomous driving,'' \emph{arXiv preprint arXiv:2311.16813}, 2023.

\bibitem{xiong2023learning}
Y.~Xiong, W.-C. Ma, J.~Wang, and R.~Urtasun, ``Learning compact representations for lidar completion and generation,'' in \emph{Proceedings of the IEEE/CVF Conference on Computer Vision and Pattern Recognition}, 2023, pp. 1074--1083.

\bibitem{lange2022lopr}
B.~Lange, M.~Itkina, and M.~J. Kochenderfer, ``Lopr: Latent occupancy prediction using generative models,'' \emph{arXiv preprint arXiv:2210.01249}, 2022.

\bibitem{zou2023diffbev}
J.~Zou, Z.~Zhu, Y.~Ye, and X.~Wang, ``Diffbev: Conditional diffusion model for bird's eye view perception,'' \emph{arXiv preprint arXiv:2303.08333}, 2023.

\bibitem{wang2024occgen}
G.~Wang, Z.~Wang, P.~Tang, J.~Zheng, X.~Ren, B.~Feng, and C.~Ma, ``Occgen: Generative multi-modal 3d occupancy prediction for autonomous driving,'' in \emph{European Conference on Computer Vision}.\hskip 1em plus 0.5em minus 0.4em\relax Springer, 2024, pp. 95--112.

\end{thebibliography}
